\documentclass[conference]{IEEEtran}
\IEEEoverridecommandlockouts
\usepackage{cite}
\usepackage{amsmath,amssymb,amsfonts}
\usepackage{algorithmic}
\usepackage{graphicx}
\usepackage{textcomp}
\usepackage{xcolor}
\usepackage{bbm}
\usepackage{float}

\usepackage{tablefootnote}

\def\BibTeX{{\rm B\kern-.05em{\sc i\kern-.025em b}\kern-.08em
    T\kern-.1667em\lower.7ex\hbox{E}\kern-.125emX}}

\title{A Knowledge-Based Decision Support System for In Vitro Fertilization Treatment
{\footnotesize }
\thanks{``*" indicates equal contribution.}}
\date{}

\author{
Xizhe Wang*$^1$, Ning Zhang*$^1$, Jia Wang*$^1$, Jing Ni$^1$, Xinzi Sun$^1$, John Zhang$^2$, Zitao Liu$^2$, Yu Cao$^1$, Benyuan Liu$^1$ \\
$^1$Department of Computer Science, University of Massachusetts Lowell  \\
$^2$New Hope Fertility Center\\
\{xizhe\_wang, ning\_zhang, jia\_wang2, jing\_ni, xinzi\_sun\}@student.uml.edu\\ 
\{john.zhang, zitao.liu\}@nhfc.com\\
\{ycao, bliu\}@cs.uml.edu
}

\begin{document}

\maketitle

\begin{abstract}
% In Vitro Fertilization (IVF) is the most widely used assisted reproductive technology. A full IVF treatment procedure usually involves ovarian stimulation, egg retrieval, in vitro fertilization, and embryo transfer. The first two steps correspond well with females' menstrual period. Therefore, we refer to it as the treatment cycle in our paper. The treatment cycle is crucial because, on the one hand, the IVF treatment protocols and medications act directly on patients' bodies so that both the clinical and side effects of medications should be considered carefully and prompt treatment adjustments are in need. On the other hand, the quality and quantity of the eggs retrieved will directly determine the results of the following procedure. To improve the success rate of treatment cycles, we propose a knowledge-based decision support system for IVF treatment that can provide medical advice on the treatments and medications for each patient visit of IVF treatment cycles. Our system is efficient in data processing and light-weighted which can be easily embedded into electronic medical record systems. Moreover, an egg retrieval oriented evaluation demonstrates that our system performs well in terms of accuracy of advice for the treatments and medications.
\textit{In Vitro} Fertilization (IVF) is the most widely used Assisted Reproductive Technology (ART). IVF usually involves controlled ovarian stimulation, oocyte retrieval, fertilization in the laboratory with subsequent embryo transfer. The first two steps correspond with females’ follicular phase and ovulation in their menstrual cycle. Therefore, we refer to it as the treatment cycle in our paper. The treatment cycle is crucial because the stimulation medications in IVF treatment are applied directly on patients. In order to optimize the stimulation effects and lower the side effects of the stimulation medications, prompt treatment adjustments are in need. In addition, the quality and quantity of the retrieved oocytes have a significant effect on the outcome of the following procedures. To improve the IVF success rate, we propose a knowledge-based decision support system that can provide medical advice on the treatment protocol and medication adjustment for each patient visit during IVF treatment cycle. Our system is efficient in data processing and light-weighted which can be easily embedded into electronic medical record systems. Moreover, an oocyte retrieval oriented evaluation demonstrates that our system performs well in terms of accuracy of advice for the protocols and medications.
\end{abstract} 


\section{Introduction}
Infertility is ranked the fifth highest severe disability for women of reproductive age in the world \cite{who2011world}. According to a World Health Organization (WHO) report, infertility affects over 50 million of couples globally \cite{who2018hrp}. Approximately 8\%-10\% surveyed couples suffer from infertility problems to some degree \cite{who2016fertility}. The occurrence rates of infertility in male and female populations are almost identical \cite{yenkie2012modeling}. Still, women suffer more from this problem as there are more and complicated causes of infertility to female patients \cite{who2016fertility}. The treatments for female patients are thus one of the major healthcare challenges worldwide.

\textit{In Vitro} Fertilization (IVF) \cite{peters1975follicular,zhang2019vitro, lu2004potential, chang1998antral, m_ivf, u_ivf, n_ivf} is one of the most widely utilized Assisted Reproductive Technology (ART). According to a national report of Centers for Disease Control and Prevention (CDC) \cite{who2016fertility}, more than $99\%$ ART cycles completed in United States employed IVF cycles in 2016. A typical IVF course consists of four stages \cite{yenkie2012modeling, zhang2019vitro}: ovarian stimulation \cite{knobil1988neuroendocrine}, oocyte retrieval \cite{lu2004potential, tan2012innovative}, fertilization in the laboratory, and embryo transfer \cite{uyar2009frequency, uyar2010bayesian}. In ovarian stimulation, medicines are cautiously applied to female patients, at the same time, testing indicators, such as hormone levels \cite{janat2009drug}, follicles \cite{peters1975follicular} and cysts \cite{chang1998antral}, are monitored. The goal of these treatments is to increase the quantity and quality of oocytes to be retrieved and the collected oocytes are fertilized in laboratory. Then the resulting embryos will be transferred into the uterus of the objective women.

Technically, the quantity and quality of the retrieved oocytes are the crucial criteria to assess  IVF treatment cycle (ovarian  stimulation and oocyte retrieval) during an entire IVF cycle. To achieve the goal of retrieving high-quality oocyte, the IVF protocol and medications should be applied cautiously since a minor clinical misjudgment or inadequate clinic attention  can cause  the IVF cycles to fail. Therefore,  IVF treatment cycle is a demanding healthcare procedure with highly involved expertise in every detail. However, the shortage of health providers, such as physicians and nurses is a critical issue in IVF institutions. As a result, the health providers are often overwhelmed by heavy workloads.

In this paper, we propose a knowledge-based IVF treatment decision support system that assists clinical judgement by providing recommendations for protocol and medication decisions during the IVF cycles. As a light-weighted design, this system can be readily integrated into existing electronic medical record (EMR) systems. With the assistance of our IVF system, the protocols and medication regimens to the IVF cycles are more efficient and accurate due to the high performance in suggestions of IVF treatment with explicit reasons provided.

Collaborating with the reproductive specialists of New Hope Fertility Center (NHFC), we devised a four-block IVF treatment protocol, which contains cycle preparation, ovarian stimulation, trigger, and post- trigger. Patient information, such as the age, menstrual cycle length, blood test and sonography scan results, is also integrated into  the system to generate the IVF treatments and medications as clinical management suggestions.

\begin{figure*}
\centering
\includegraphics[width=180mm]{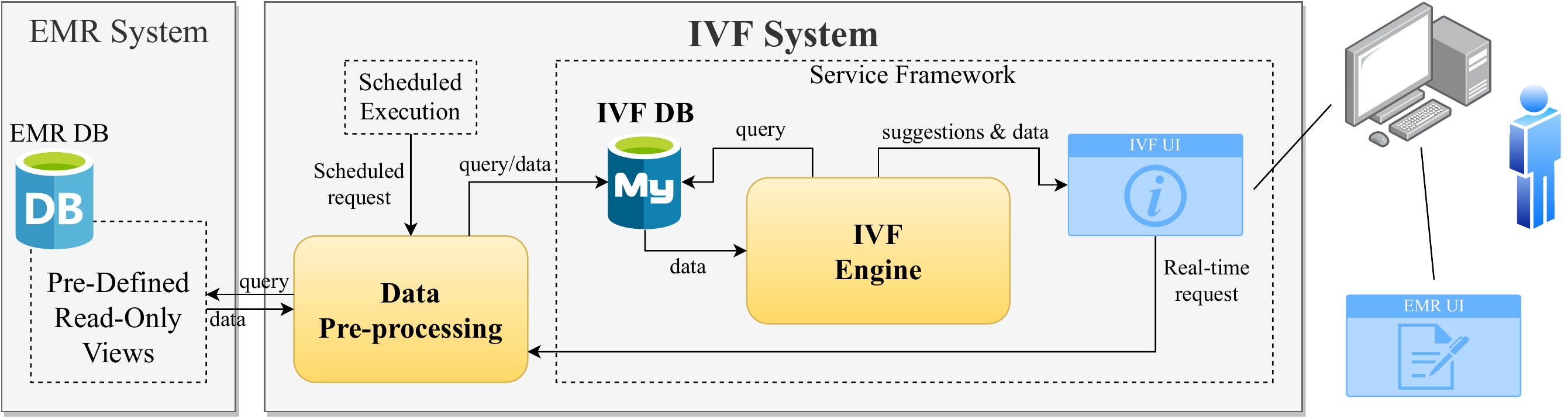}
\caption{System Architecture.}
\label{system_architechture}
\end{figure*}

There are more than 8,000 female patients’ data with over 20,000 IVF cycles collected in NHFC from December 2011 to March 2020. So far, we obtained 16,513 cycles  with  99,577 records available for assessment after data cleaning. The evaluation of the IVF system shows that the accuracy of decision suggestions is 83.95\% for block 1, 79.59\% for block 2, 89.60\% for block 3, and 95.42\% for block 4. Particularly, the accuracy is 75.10\% for turning point from block 1 to block 2, 94.70\% for turning point from block 2 to block 3, 83.62\% for turning point from block 3 to block 4, and 73.13\% for turning point from block 4 to Luteal Phase Stimulation (LPS) \cite{lps}.

\section{Related Work}

There are also other studies that focus on particular stages in IVF treatments. Liu and Hu \cite{lu2004potential} introduced a potential assistant robot for IVF oocyte retrieval. Tan \textit{et al.} \cite{tan2012innovative} proposed an integrated prototype for oocyte retrieval stage. Moreover, Uyar \textit{et al.} \cite{uyar2009frequency} proposed a frequency based encoding technique and SVM to perform the embryo classification. Later, paper \cite{uyar2010bayesian} presented Bayesian networks to classify embryos.

The brief interpretations for several frequently used hormone terms are listed as follows.

\begin{itemize}

\item FSH \cite{fsh}: Follicle Stimulating Hormone is created by the pituitary gland. It regulates the ovary functions.

\item LH \cite{lh}: Luteinizing Hormone (lutropin/lutrophin) is produced by the gonadotropic cells. An acute rise of LH will trigger ovulation and development of the corpus luteum.

\item E2 \cite{e2}: Estrodial is one of sex hormones that develops and maintains the female characteristics.

\item P4 \cite{p4}: Progesterone promotes the preparation of the endometrium for the potential pregnancy after ovulation.

\item HCG \cite{hcg}: Human Chorionic Gonadotropin (HCG) is a chemical produced by trophoblast. It is helpful in identifying the pregnancy quality.

\end{itemize}

\section{System Architecture}
The knowledge-based IVF treatment decision support system is constructed of three top-level modules: an IVF treatment decision support engine, a data pre-processing module, and an IVF database. An overview of the system architecture is shown in Figure \ref{system_architechture}. This light-weighted system can be readily integrated into an EMR system with two interfaces, namely, several read-only views offered from EMR database and a web request from an EMR User Interface (UI) operation.

In practice, a physician opens EMR UI to access patient records, and then further obtains the IVF treatment suggestions from our proposed IVF system. Upon requests, the data pre-processing module downloads data from pre-defined views in EMR database and reorganizes them into the format our system requires. Then we store the data into our IVF database. The data pre-processing runs both in scheduled tasks for daily data and on-demand tasks for real time requests. After being activated by requests, the IVF engine retrieves data from the IVF database, and generates corresponding advice and suggestions for IVF treatments and medications. Finally, the IVF system creates a pop-up webpage with the patient profile and treatment suggestions.

\subsection{IVF Decision Support Engine}
We divide an IVF cycle into four blocks, namely, preparation, stimulation, trigger, and post-trigger. In each of these blocks, there are several tests (blood tests and ultrasound tests) and decisions for treatments. The process of the IVF engine can be structured as work flow mechanisms with these four blocks. When it comes to the clinical diagnosis of a patient, the IVF engine will give treatment suggestions based on her current blood test, ultrasound test as well as the status she was in before the clinic visit.

\begin{figure}[H]
\centering
\includegraphics[width=85mm]{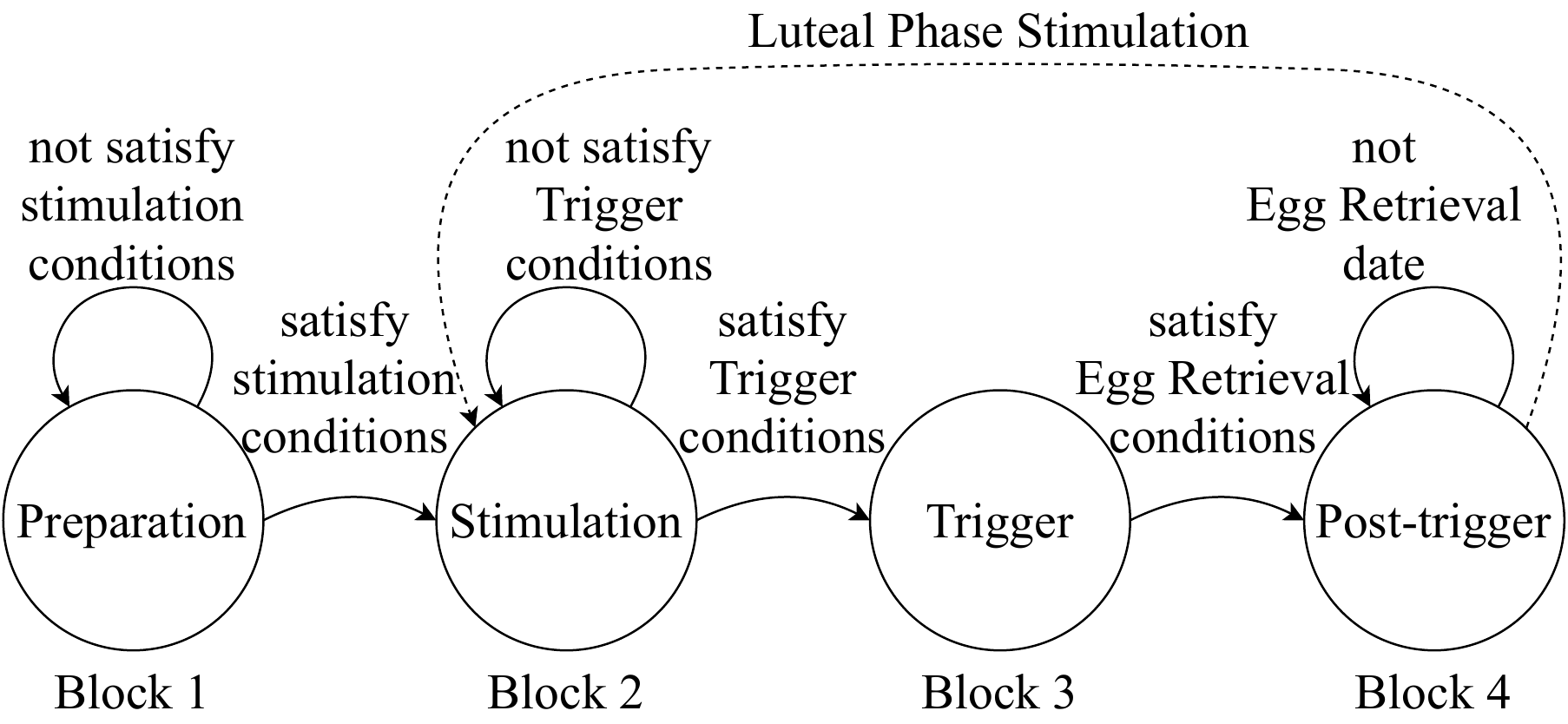}
\caption{Four-Block IVF Cycle Protocol.}
\label{std_4_block}
\end{figure}

An IVF cycle can be structured as a four-block protocol. Figure \ref{std_4_block} depicts a concise finite state machine diagram of such protocol. Occasionally, the Lutreal Phase Stimulation (LPS) protocol might be applied due to patient circumstances \cite{lps}. A significant difference between the LPS protocol and the standard stimulation protocol (Follicular Phase Stimulation protocol) is that the stimulation process of the LPS protocol starts after the oocyte retrieval procedure of the standard stimulation protocol, which is denoted as the arrow with dashed curve in Figure \ref{std_4_block}.

\noindent \textit{\textbf{1) Block 1: Preparation}}.

\begin{figure}[!ht]
\centering
\includegraphics[width=81mm]{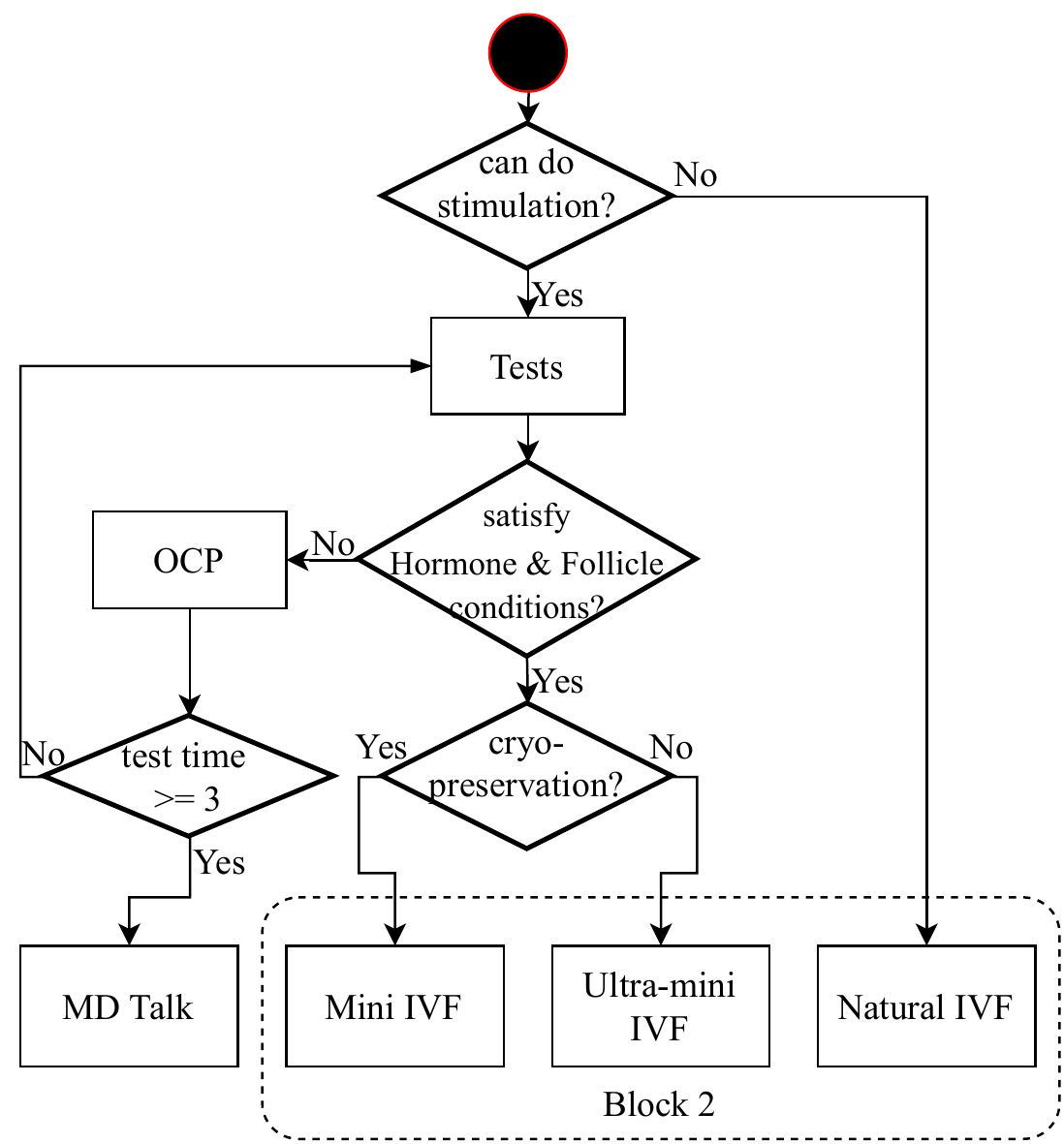}
\caption{Block 1 Protocol.}
\label{block_1_fc}
\end{figure}

In preparation block, patients generally take birth control pills until their hormone level and antral follicle reserve are good for the upcoming IVF cycle. Figure \ref{block_1_fc} shows the work flow of block 1. The hormone and follicle conditions of block 1, shown in Table \ref{b1_HF_cond}, are the crucial criteria to determine the block starting time. The criteria can be divided into two parts by patient age since their hormone circumstances are highly related to ages. Generally, the test interval varies from five days to nine days. After the end of block 1, an explicit stimulation scheme will be proposed as the start of block 2.

\begin{table}[!ht]
\centering
\caption{Block-1 Hormone and Follicle Conditions}
\begin{tabular}{c|cc}
\hline
 & age $< 42$ & $42 \leq$ age  \\ \hline
FSH & $< 15$ & $< 15$ \\ \hline
LH & $< 8.5$ & $< 6$ \\ \hline
E2 & $< 50$ & $< 65$ \\ \hline
P4 & $< 1.5$ & $< 1.5$ \\ \hline
Number of Follicle & $\geq 45 - age$ & $1\sim 6$ \\ \hline
Follicle Size & $\leq8$mm & $\leq8$mm\\ \hline
\end{tabular}
\label{b1_HF_cond}
\end{table}

There are several possible treatments that the IVF engine might propose in block 1.
\begin{itemize}

\item OCP (Oral Contraceptive Pills): It implies that the hormone level and antral follicle reservation are not good enough to start stimulation so that the patient should continue to take OCP.

\item MD Talk: Doctors meet with the patients and talk about their situations. It occurs mainly because the patients show poor reactions to the IVF treatments.

\end{itemize}

\noindent \textit{\textbf{2) Block 2: Stimulation}}.

%% The Edition Before Removing Medications
%In this block, a specific stimulation scheme is proposed at the start, and the patient will continuously receive stimulation treatments while the medications and dosages can vary over time. The work flow of stimulation protocol is illustrated in Figure \ref{block_2_1_fc}. The medication protocols, with regard to both first visit case and non-first visit case in stimulation stage, are illustrated in Figure \ref{block_2_2_1st_v_fc} and Figure \ref{block_2_2_not_1v_fc}, respectively. Moreover,  the processes and medications in stimulation block are determined by the IVF scheme decided at the end of block 1. Usually, Mini IVF scheme and Ultra-mini IVF scheme share similar protocols, while Natural IVF scheme works in a relatively simpler way.

\begin{figure}[!ht]
\centering
\includegraphics[width=85mm]{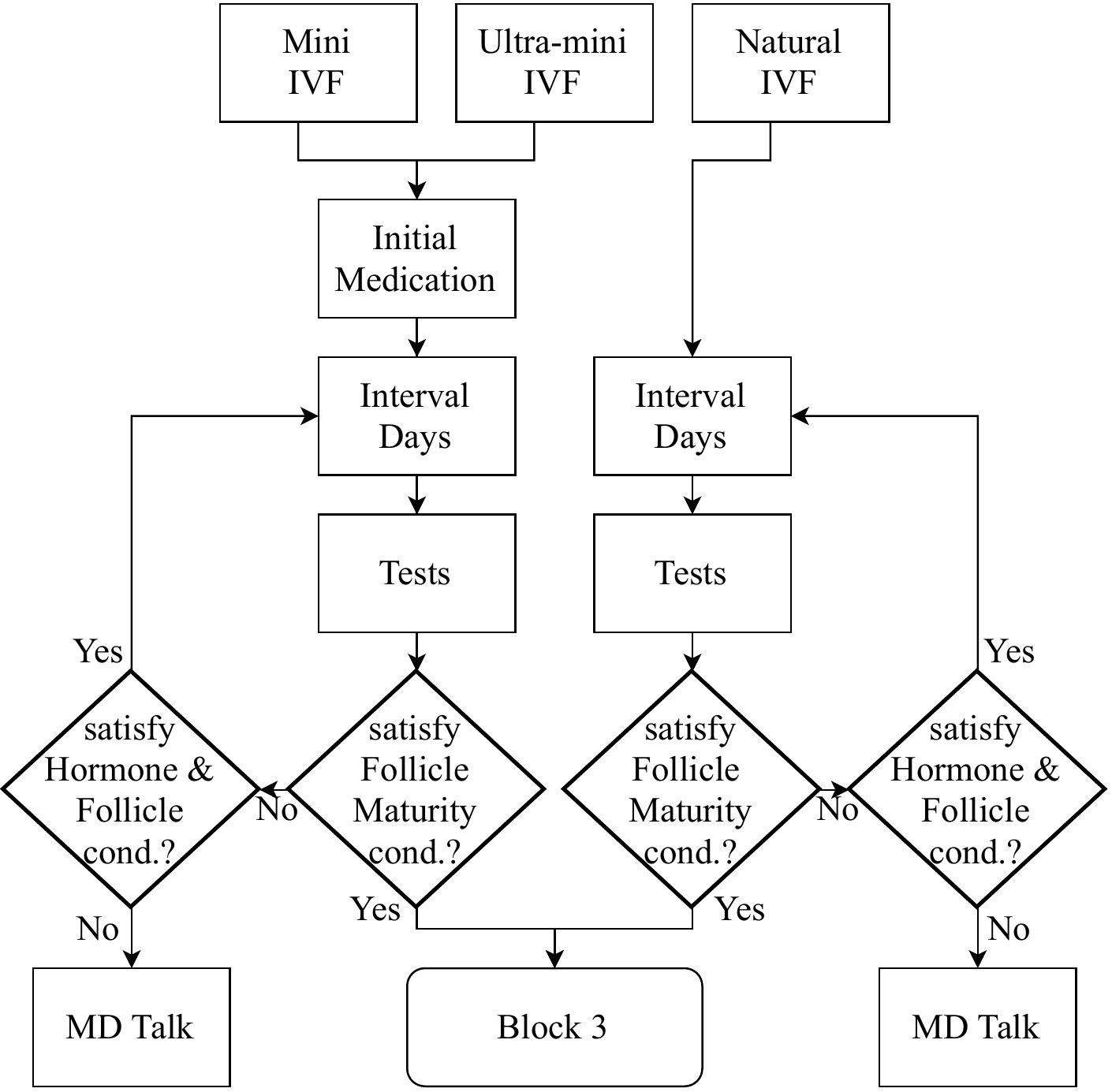}
\caption{Block 2 Protocol.}
\label{block_2_1_fc}
\end{figure}

In this block, a specific stimulation scheme is proposed at the start, and the patient will continuously receive stimulation treatments while the medications and dosages can vary over time. The work flow of stimulation protocol is illustrated in Figure \ref{block_2_1_fc}. Moreover,  the processes and medications in stimulation block are determined by the IVF scheme decided at the end of block 1. Usually, Mini IVF scheme and Ultra-mini IVF scheme share similar protocols, while Natural IVF scheme works in a relatively simpler way.

%\begin{figure}[!ht]
%\centering
%\includegraphics[width=80mm]{Figs/block_2_2_1st_v_fc.pdf}
%\caption{Medication Protocol, first clinic visit in Block 2.}
%\label{block_2_2_1st_v_fc}
%\end{figure}

%\begin{figure}[!ht]
%\centering
%\includegraphics[width=80mm]{Figs/block_2_2_not_1v_fc.pdf}
%\caption{Medication Protocol, non-first clinic visit in Block 2.}
%\label{block_2_2_not_1v_fc}
%\end{figure}

The purposes of stimulation are to stimulate the growth of follicles and to find the right timing to trigger. The follicle maturity conditions, listed in Table \ref{b2_FM_cond}, are the rules to determine if follicles have been mature enough such that the IVF cycle can be transitioned into the next block. The hormone and follicle conditions, listed in Table \ref{b2_HF_cond}, are used for evaluating the follicle growth in the cycles. The length of this block is relatively stable and corresponds to the follicular phase of the patient. Normally, a patient will receive stimulation treatments for less than two weeks. In each clinic visit, the protocol also allows to adjust the medications and dosages by a patient's physical reactions. The IVF engine suggests ``Continue Stimulation" during block 2.

\begin{table}[!ht]
\centering
\caption{Block-2 Follicle Maturity Conditions}
\begin{tabular}{c|cc}
\hline
 & $15$mm & $18$mm  \\ \hline
Follicle Percentage & $\geq 60\%$ & $\geq 30\%$ \\ \hline
\end{tabular}
\label{b2_FM_cond}
\end{table}

\begin{table}[!ht]
\centering
\caption{Block-2 Hormone and Follicle Conditions}
\begin{tabular}{c|cc}
\hline
& Mini IVF/Ultra-mini IVF & Natural IVF \\ \hline
FSH & $15\sim25$ & $5\sim25$ \\ \hline
LH & $< 15$ & $2\sim15$ \\ \hline
E2 & $> 50$ & $> 80$ \\ \hline
P4 & $< 1.2$ & $< 1$ \\ \hline
Follicles & growing & growing \\ \hline
\end{tabular}
\label{b2_HF_cond}
\end{table}

In fact, it is a frequent phenomenon that patients' follicles grow slowly even with the help of medications. In such cases, the IVF engine adjusts the medications and dosages, or even changes the stimulation scheme due to the hormone/follicle indicators obtained from tests. Meanwhile, it also memorizes that if the patient has a poor response to certain medicines. When a patient's follicles have grown to certain sizes, e.g. $60\%$ of antral follicles are larger than $15$mm or $30\%$ of antral follicles are larger than $18$mm, she will proceed to block 3.

There are several different IVF treatments that the IVF engine might propose and some terms in block 2.
\begin{itemize}

\item Mini IVF \cite{m_ivf}: An IVF stimulation scheme that uses weak medications or low dosages of medications, compared to conventional IVF protocols.

\item Ultra-mini IVF \cite{u_ivf}: An IVF stimulation scheme that uses weaker medications or lower dosages of medications than the Mini IVF.

\item Natural IVF \cite{n_ivf}: An IVF stimulation scheme that no medication is used at all. It is adopted for patients whose bodies are not suitable for the IVF stimulation medications.

\item Initial Medication: Follism/Gonal F: $150$IU; Clomid: $50$mg; Letrozole: $2.5$mg.

\item Interval Days: Typically, the pattern of interval days are $5$, $3$, $1$, $1$, $1$. In the beginning of block 2, the IVF engine proposes an initial medication to be taken for $5$ days. After $5$ days, the patient returns to the clinic for further tests and re-evaluation. The updated medication treatment lasts $3$ days, then the patient will return for tests and treatment update. The process repeats according to the interval pattern. These steps will not stop until a block-change decision or MD Talk decision is made. In practice, the pattern of interval days might be slightly amended by the doctors due to patients' special circumstances.

%\item Continue Stimulation: The IVF scheme decided at the end of block 1 significantly impacts the choices of medication treatments in block 2. In each clinic visit, the protocol also allows to adjust the medications and dosages by a patient's physical reactions. The IVF engine suggests ``Continue Stimulation" during block 2.

%\item Standard Medicines: Clomid: $50$mg; Follism: $150$IU.

%\item Half Medicines: Clomid: $25$mg; Follism: $75$IU.

% \item afs: antral follicles.

\end{itemize}

\noindent \textit{\textbf{3) Block 3: Trigger}}.

In this block, patients receive ``trigger shot" with trigger medications for a trigger duration. The effect of the ``trigger shot" is to develop oocytes into meiosis where the number of chromosomes is cut in half (from 46 to 23). The work flow of trigger protocol is illustrated in Figure \ref{block_3_fc}. When a patient finishes tests, the engine will first determine trigger duration by LH level and patient age. In Figure \ref{block_3_fc}, ``prev" represents the LH level obtained from previous blood test. The increase rate of LH is a key indicator to trigger duration. If LH $>$ 25, a patient is triggered with no medication, and the engine will determine the trigger medications based on E2 level and follicle grow circumstances. If E2 $<$ 4000, the trigger medication will be Lupron (1 unit). Otherwise, if the number of follicles larger than 15mm is less than 6, the trigger medication will be Lupron (1 unit) or Ovidrel. If the number equals to or is larger than 6, we trigger with Lupron (2 units) or Lupron (1 unit) + Ovidrel.

There are several different treatments and medications that the IVF engine might propose in block 3.
\begin{itemize}

\item Trigger Duration: The duration between ``Trigger" and ``Oocyte Retrieval".

\item No Trigger: A trigger scheme that no medication is applied and it should be determined less than 48 hours before oocyte retrieval.

\item Lupron: It overstimulates the body's own production of certain hormones, causing that production to shut down temporarily. It reduces the amount of testosterone in men or estrogen in women. The number after ``Lupron" indicates the dosage unit applied.

\item Ovidrel: Ovidrel is the brand name for the hormone fertility drug chorionic gonadotropin alpha. It is r-hCG or recombinant human chorionic gonadotropin.

\begin{figure}[!ht]
\centering
\includegraphics[width=72mm]{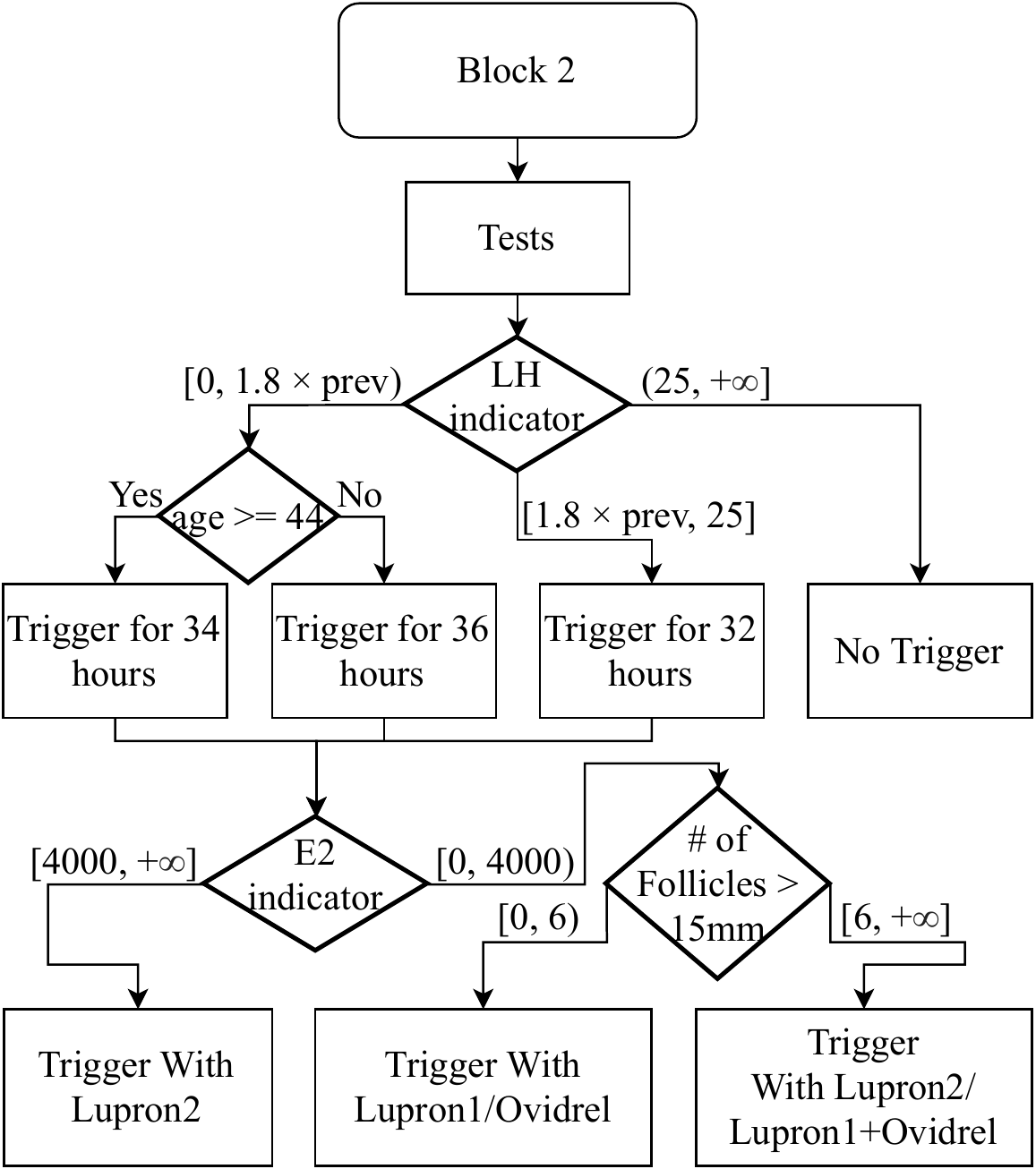}
\caption{Block 3 Protocol.}
\label{block_3_fc}
\end{figure}

\end{itemize}

\noindent \textit{\textbf{4) Block 4: Post-Trigger}}.

\begin{figure}[!ht]
\centering
\includegraphics[width=38mm]{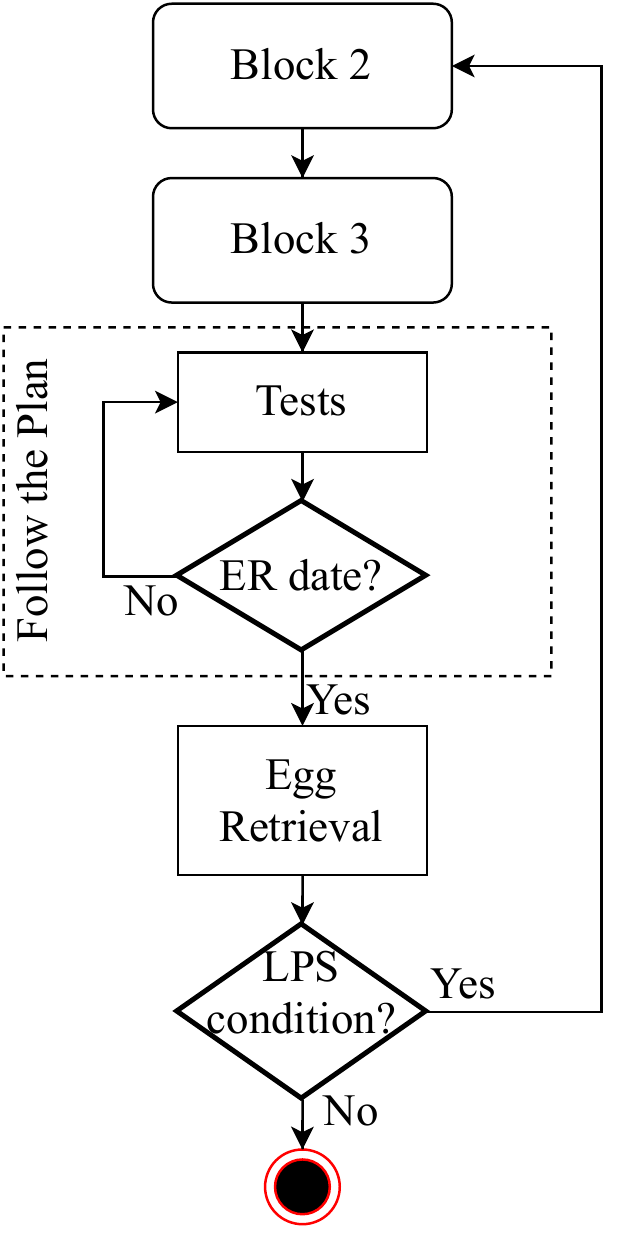}
\caption{Block 4 Protocol.}
\label{block_4_fc}
\end{figure}

There are two major steps in post-trigger: follow the plan and oocyte retrieval. The work flow of block 4 protocol is illustrated in Figure \ref{block_4_fc}. After oocyte retrieval is done, the current IVF cycle is considered to be finished. However, if the patient also take the Luteal Phase Stimulation, the IVF cycle would continue. The time span from the beginning of Trigger to the end of Post-Trigger should be less than 2 days. Otherwise, ovulation might happen, which leads to failure for oocyte retrieval.

There are several different treatments that the IVF engine might propose in block 4.
\begin{itemize}

\item Follow the Plan: Stick to the proposed trigger scheme, including the medications and scheduled oocyte retrieval date-time.

\item LPS conditions: The age of patient equals to or larger than 40, and the number of follicles smaller than 18mm is larger than 4. LPS is a special operation that an IVF stimulation is carried out in luteal phase. LPS is used because ovulation function is relatively weak for age advanced female patients such that the protocol tends to make full use of every chance when there are potentially available follicles.

\end{itemize}

\subsection{Data Pre-processing}
\begin{figure}[!ht]
\centering
\includegraphics[width=85mm]{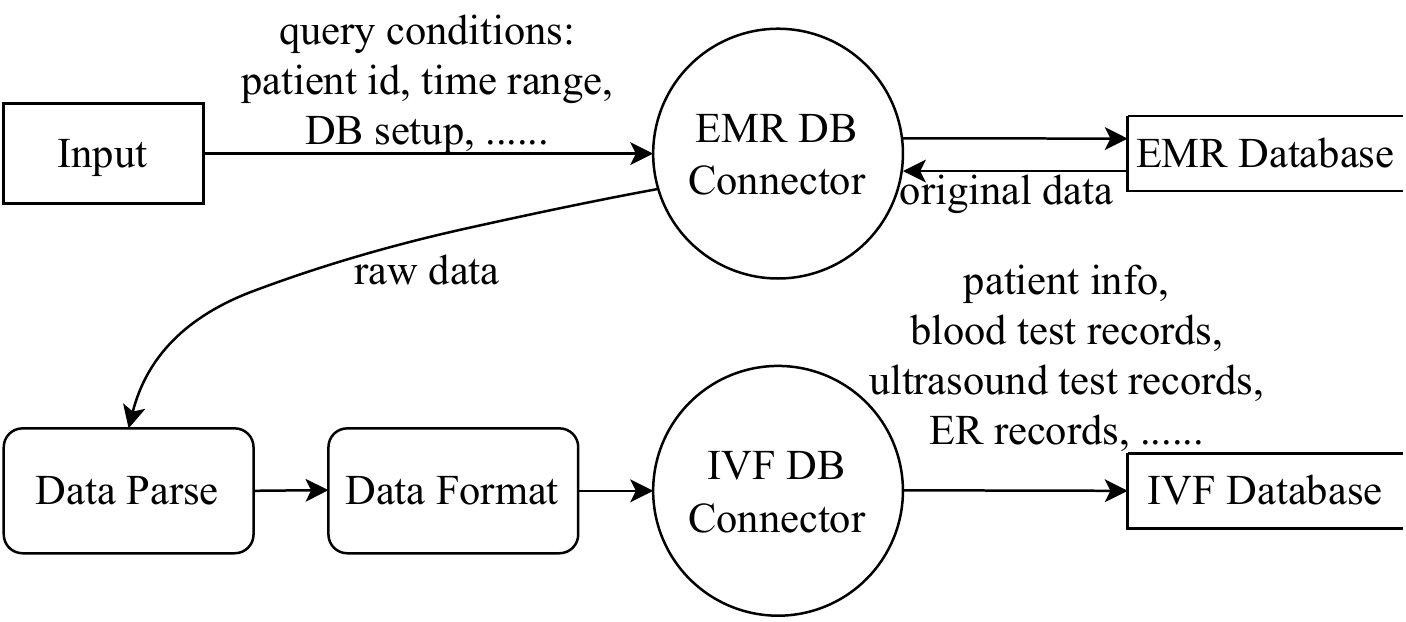}
\caption{Level-1 Data Flow Diagram of Data Pre-processing Module.}
\label{data_pre_pro}
\end{figure}

Data pre-processing module provides the IVF engine with patient information and test data. Figure \ref{data_pre_pro} illustrates a level-1 data flow diagram of data pre-processing. It is consisted of four basic sub-modules: data parsing, data formatting, and two database connectors.

\noindent \textit{1) Data Parsing}.
For blood tests, ultrasound tests, and oocyte retrievals, all the data, including the hormone levels, follicle and oocyte retrieval statistics, should be stored into IVF database as integer or float values. However, the original data obtained from the record tables of EMR database cannot be directly used because there are many inputs and spelling exceptions. This module aims at filtering such exceptions and providing the data to next steps.

\noindent \textit{2) Data Formatting}.
Data formatting sub-module works on formatting the integer/float data output from data parsing sub-module so that the data can be stored into IVF database. There are five IVF-related tables in IVF database that need the update from data formatting sub-module. They will be discussed in sub-section \textit{C}.

The processes of formatting data for patient information, blood tests, and oocyte retrievals are relatively straightforward as all the parsed integer/float data will be directly stored into the matched columns of appointed tables. Formatting data for ultrasound tests requires an extra operation of mapping the sizes and numbers of follicles. The numerical representation of the follicle information in an ultrasound test is organized as a json data-structure.

\noindent \textit{3) Database Connectors}.
The Data Pre-processing module contains two database connectors: an EMR database connector and an IVF database connector.

\begin{itemize}

\item EMR database connector downloads data from several read-only views pre-defined in EMR database. Original data from EMR database are returned to the data parsing sub-module. This connector is implemented with a MS SQL Server connection plugin since the EMR database in NHFC is implemented as a MS SQL Server.

\item IVF database connector can download and upload data. This connector is implemented with a MySQL connection plugin as the IVF database is constructed with open source MySQL.

\end{itemize}

%\noindent \textit{4) Two Tasks for Data Pre-processing}.
%There are two tasks that utilize the data pre-processing module: scheduled task and real-time task. The time sequence diagram is shown in Fig. \ref{two_tasks}.

%\begin{figure}[!ht]
%\centering
%\includegraphics[width=85mm]{Figs/two_tasks.pdf}
%\caption{Time Sequence Diagram of Data Pre-processing.}
%\label{two_tasks}
%\end{figure}

%\begin{itemize}

%\item Scheduled Task. This task automatically calls the data pre-processing module at 1:00 a.m. every days. It downloads all the patient data from EMR database updated in the previous day. The data will be processed and stored into IVF Database as daily increment without altering existing data.

%\item On-demand Task. This task utilizes the data pre-processing module when a doctor requests the assistance of our IVF system at EMR UI. When received a request referring to a given patient, the data pre-processing module will query and download the specific patient's data from EMR database of current day. Moreover, data pre-processing module can be called through real-time task any time during a day. A patient's new data will update its existing version if these two versions of data have the same time stamp.

%\end{itemize}

\subsection{Database}
The IVF database is constructed using MySQL Community (GPL). The database is built on a server that connects to the local area network in NHFC, and only provides access to users within the LAN. The tables in IVF database can be divided to two parts based on their purposes: IVF-related tables and Django supporting tables.

\begin{figure}[!ht]
\centering
\includegraphics[width=88mm]{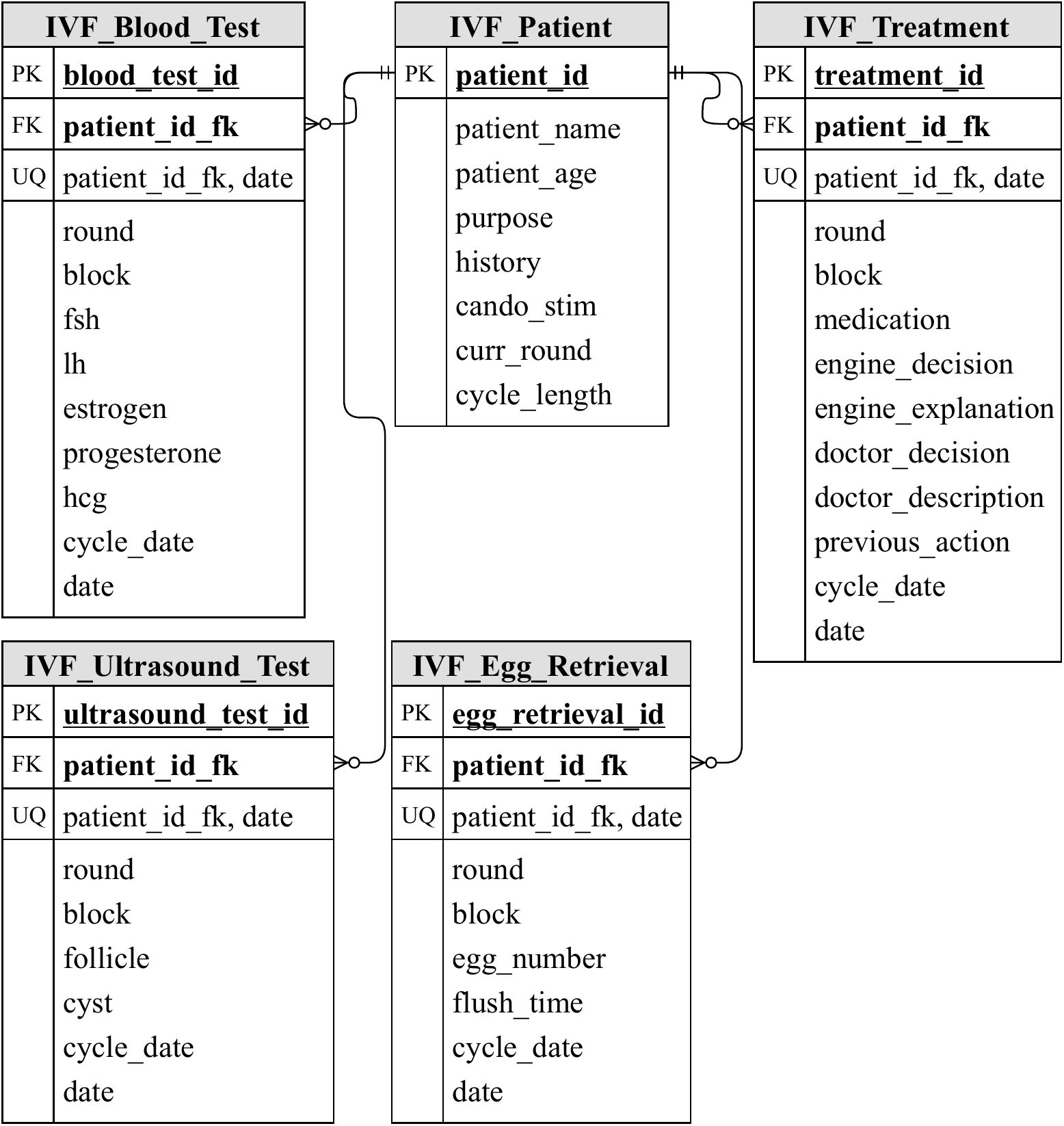}
\caption{Entity-Relationship Diagram of IVF-related Tables.}
\label{ivf_db_er}
\end{figure}

There are five IVF-related tables including \textit{IVF\_Patient}, \textit{IVF\_Blood\_Test}, \textit{IVF\_Ultrasound\_Test}, \textit{IVF\_Egg\_Retrieval} and \textit{IVF\_Treatment}. Figure \ref{ivf_db_er} shows the entity-relationship diagram of the tables. IVF\_Patient is a restriction table that the other four IVF-related tables must comply with the restrict of attribute patient\_id as a foreign key.

%\begin{figure}[!ht]
%\centering
%\includegraphics[width=88mm]{Figs/db_big_pic.pdf}
%\caption{Data Flow of IVF-related Tables.}
%\label{db_big_pic}
%\end{figure}

%Among the IVF-related tables, the main purpose of \textit{IVF\_Patient}, \textit{IVF\_Blood\_Test}, \textit{IVF\_Ultrasound\_Test}, and  \textit{IVF\_Egg\_Retrieval} is to provide data, such as patient information, test results, and egg statistics, to the IVF Engine. The IVF treatment advice generated by the engine is returned and stored to table \textit{IVF\_Treatment}. Besides, an attribute ``block" is also updated by the IVF engine in each table. The data flow between the tables and the IVF engine is illustrated in Fig. \ref{db_big_pic}.

The Django supporting tables are built to enable the Django service as the integration framework. There are tables for users, login authorities, sessions and so on.

\section{Implementation and Evaluation}
\subsection{System Implementation}
\begin{figure*}
\centering
\includegraphics[width=160mm]{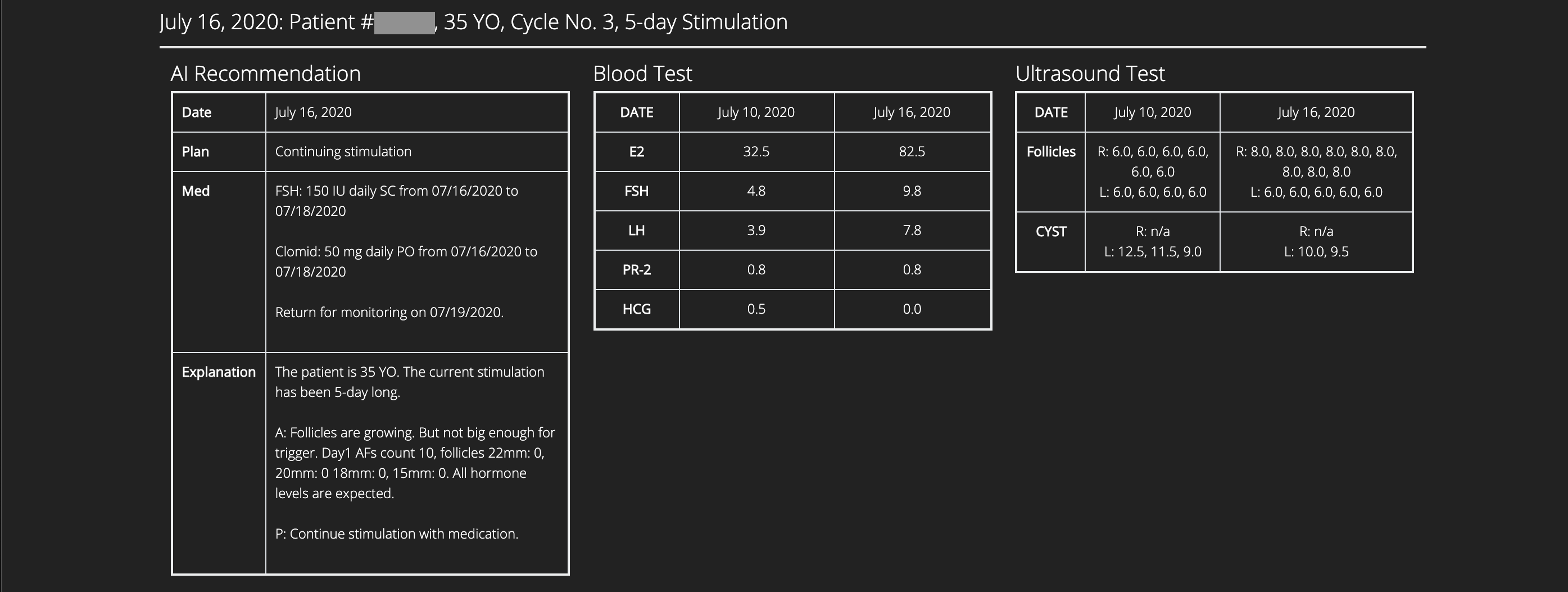}
\caption{The User Interface of the IVF Engine}
\label{ui}
\end{figure*}

The knowledge-based assistant system is implemented on a desktop with Ubuntu 16.04 Operating System. On the back-end side, The IVF engine and data pre-processing module are developed in Python of version 3.7 with pyknow package supported as the knowledge engine resource. The IVF database is built using open resource MySQL Community. Both database connectors are implemented in the same Python environment and can connect to MS SQL Server database and MySQL database. We use Django framework to integrate every module together as the IVF system. On the front-end side, we use Django framework to develop the user interface, shown as Figure \ref{ui}, to provide the following information: i) patient basic information, such as cycle number, age, and patient ID, ii) intra-day blood and ultrasound tests results, iii) corresponding predictions of assessments and explanations, iv) corresponding predictions of medication dosage prescription.

\subsection{Performance Evaluations}
From December 2011 to March 2020, there are 142,427 pairs of blood test and ultrasound test records and 14,926 oocyte retrieval records within 20,486 IVF cycles of 8,300 patients totally in the EMR database. Our data pre-processing module parsed 99,577 IVF treatment decision records within 16,513 IVF cycles of 6,061 patients as the evaluation dataset.

% Specifically, each blood test record contains five types of hormones--Estrogen, FSH, LH, Progesterone, and HCG; each ultrasound test record contains the diameter of the existing follicles; each oocyte retrieval record contains the numbers of retrieved oocyte and flush time.

% These records are from ???? cycles of ???? patients. Table \ref{generall_statistics} shows more details about the statistics of the dataset.

% \begin{table}[!ht]
% \centering
% \caption{General Statistics of Dataset}
% \begin{tabular}{c|ccc}
% \hline
% \hline
%   & Patients & IVF Cycles & Test Records \\ \hline
% Max (one patient) & / & 17 & 67 \\ \hline
% Average (per patient) & / & 2 & 9 \\ \hline
% Total & 9093 & 12565 & 88792 \\ \hline
% \end{tabular}
% \label{generall_statistics}
% \end{table}

We use the following strategy to evaluate our IVF engine. First, we split the dataset into sub groups each of which contains data of one cycle of one patient, and the data of each sub group is sorted by date. Second,  we reset the IVF engine at the beginning of each cycle, and then feed the data of the corresponding sub group to the IVF engine, one pair after another. According to each pair of input data, The IVF engine outputs a set of predictions, which include a decision, a corresponding explanation and prescription. A decision contains information such as this patient should stay at the current block or move to the next block, or an alert if this patient has an urgent situation that needs a MD talk. Third, we evaluate the predicting decision for each pair of input data with the corresponding doctors' decision, If the predicting decision is wrong, we modify the predicting decision with the doctors' decision to avoid incorrect treatment history for subsequent predictions.

\begin{figure}[H]
\centering
\includegraphics[width=78mm]{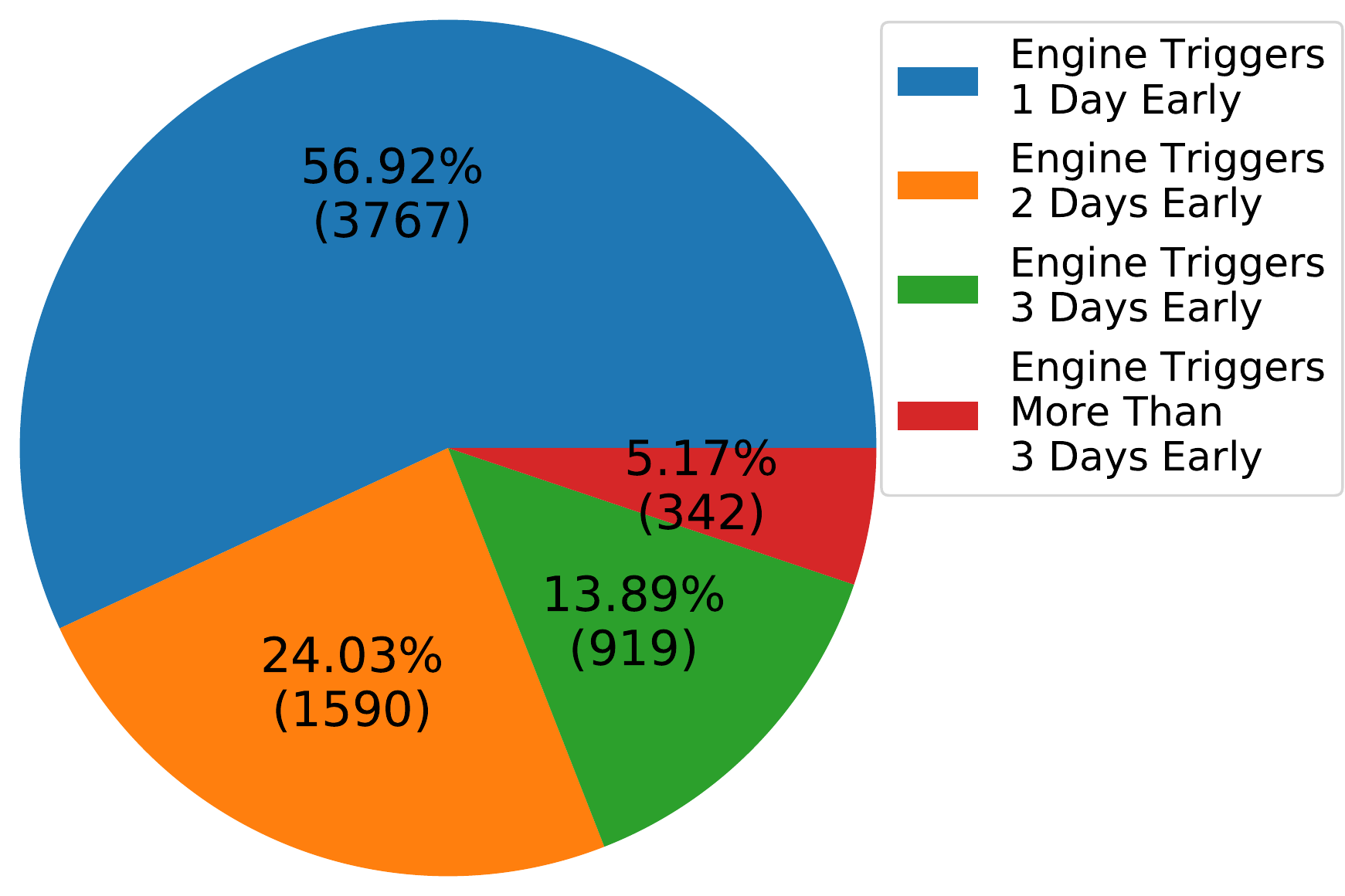}
\caption{Statistics of Early Triggers of the IVF Engine}
\label{early_trigger}
\end{figure}

% Table \ref{accuracy_1} shows the overall 71.02\%, 66.35\%, 80.16\%, and 73.46\% \footnote{We do not count the LPS precedure in the regular block 4 evaluation.}
% , however, the number of data pairs of block 3 and block 4 is approximately equal to the number of cycles
Table \ref{accuracy_1} shows 83.95\%, 79.59\%, 89.60\%, and 95.42\% intra-block accuracy for block 1 to block 4, respectively. Generally block 1 and block 2 each takes two weeks, block 3 takes 24-36 hours, and block 4 takes even shorter. Meanwhile, not all cycles can go to the oocyte retrieval stage successfully, thus the number of data pairs of block 1 and block 2 is two to three times of the number of IVF cycles. Similar to the actual clinical experience, the IVF engine has higher error rates on block 1 and block 2 compared to the latter blocks, since patients usually have more complex situations at the OCP preparation and stimulation stage. Take the trigger protocol day for instance. On the one hand, ovulations before oocyte retrieval are serious problems for IVF processes. As the virtual assistant for physicians and nurses, the IVF engine should alert immediately once blood and ultrasound test results indicate trigger requirements have been met. On the other hand, patients generally have one-to-three-day trigger window so that physicians can flexibly determine the trigger time and the subsequent oocyte retrieval time according to schedules of both patients and clinics. The aforementioned two factors result in that the IVF engine tends to predict trigger ``early'' (the B2 errors in Table VII) rather than ``late'' (the B2-B3 errors in Table VIII) as shown in TABLE VII.  Figure 13 provides more details about the B2 errors shown in TABLE VII. 56.92\% of B2 errors (3,767 over 6,618) are one-day-earlier, which implies that some patients were triggered one day after the trigger window opened. As time goes on, the risk of ovulations increases, so the percentage of more-than-one-day-earlier trigger predictions gets smaller, as shown in Figure \ref{early_trigger}.

Due to the complexity and significance of the first two blocks for the IVF protocol, we also analyzed the performance of the IVF engine for the turning points between adjacent blocks, shown in Table \ref{accuracy_2}. The IVF engine achieves an accuracy of 75.10\%, 94.70\%, 83.62\%, and 73.13\% for each transition. The reason why the B1-B2 transition has a 19.60\% lower accuracy compared to the B2-B3 transition is that, block 1 is actually a preparation process that adjusts the condition of patients from disordered hormone level or insufficient follicle reserve to the appropriate condition. However, the preparation process does not work well for all patients, and thus doctors sometimes slightly relax the threshold of the B1-B2 transition to avoid patients missing this cycle. The ``late-entry" errors include follicle size is too big to enter stimulations, or some hormone levels, such as FSH, E2, and LH, is higher than thresholds, which verifies that the B1-B2 errors mainly result from the relaxations by doctors.

The column \textit{B4-LPS} of Table \ref{accuracy_2} shows the performance of the IVF engine for Luteal Phase Stimulation (LPS) procedures. The IVF engine can correctly predict 73.13\% of the cycles that need a LPS procedure. The reasons why the IVF engine can not predict more accurately are listed as follows. Sometimes patients hope to get more oocytes retrieved in one cycle, even though their situations do not meet the standard of LPS procedures. On the other hand, since LPS procedures is just an option for patients rather than a mandatory procedure, many patients chose not to take this procedure even though their conditions satisfy the requirements of  LPS procedures.

\begin{table}[!ht]
\centering
\caption{Accuracy of Intra-block decisions}
\begin{tabular}{c|cccc}
\hline
\hline
  & B1 & B2 & B3 & B4 \\ \hline
Wrong & 4439 & 6618 & 1918 & 815 \\
Correct & 23221 & 25807 & 16513 & 17003 \\
Total & 27660 & 32325 & 18431 & 17818 \\ \hline
Accuracy & 83.95\% & 79.59\% & 89.60\% & 95.42\% \\ \hline
\end{tabular}
\label{accuracy_1}
\end{table}

% Still working on this table
% by Xizhe
\begin{table}[!ht]
\centering
\caption{Accuracy of Block transitions}
\begin{tabular}{c|ccccc}
\hline
\hline
transition & B1-B2 & B2-B3 & B3-B4 & B4-LPS \\ \hline
Wrong & 3913 & 850 & 2624 & 237 \\
Correct & 11806 & 15175 & 13401 & 645 \\
Total & 15719 & 16025 & 16025 & 882 \\ \hline
Accuracy & 75.10\% & 94.70\% & 83.62\% & 73.13\% \\ \hline
\end{tabular}
\label{accuracy_2}
\end{table}
% \footnotetext[1]{Ground truth record number of ``OCP" in Block 1.}
% \footnotetext[2]{Ground truth record number of ``IVF" in Block 2.}
% \footnotetext[3]{Ground truth record number of ``Continue Stimulation" in Block 2.}
% \footnotetext[4]{Ground truth record number of ``Trigger" in Block 3.}
% \footnotetext[5]{Ground truth record number of ``Follow the Plan" in Block 4.}
% \footnotetext[6]{Ground truth record number of ``Terminate" in Block 4.}
% \footnotetext[7]{Ground truth record number of ``LPS" in Block 4.}
% % \footnotetext[]{}

\begin{figure}[H]
\centering
\includegraphics[width=78mm]{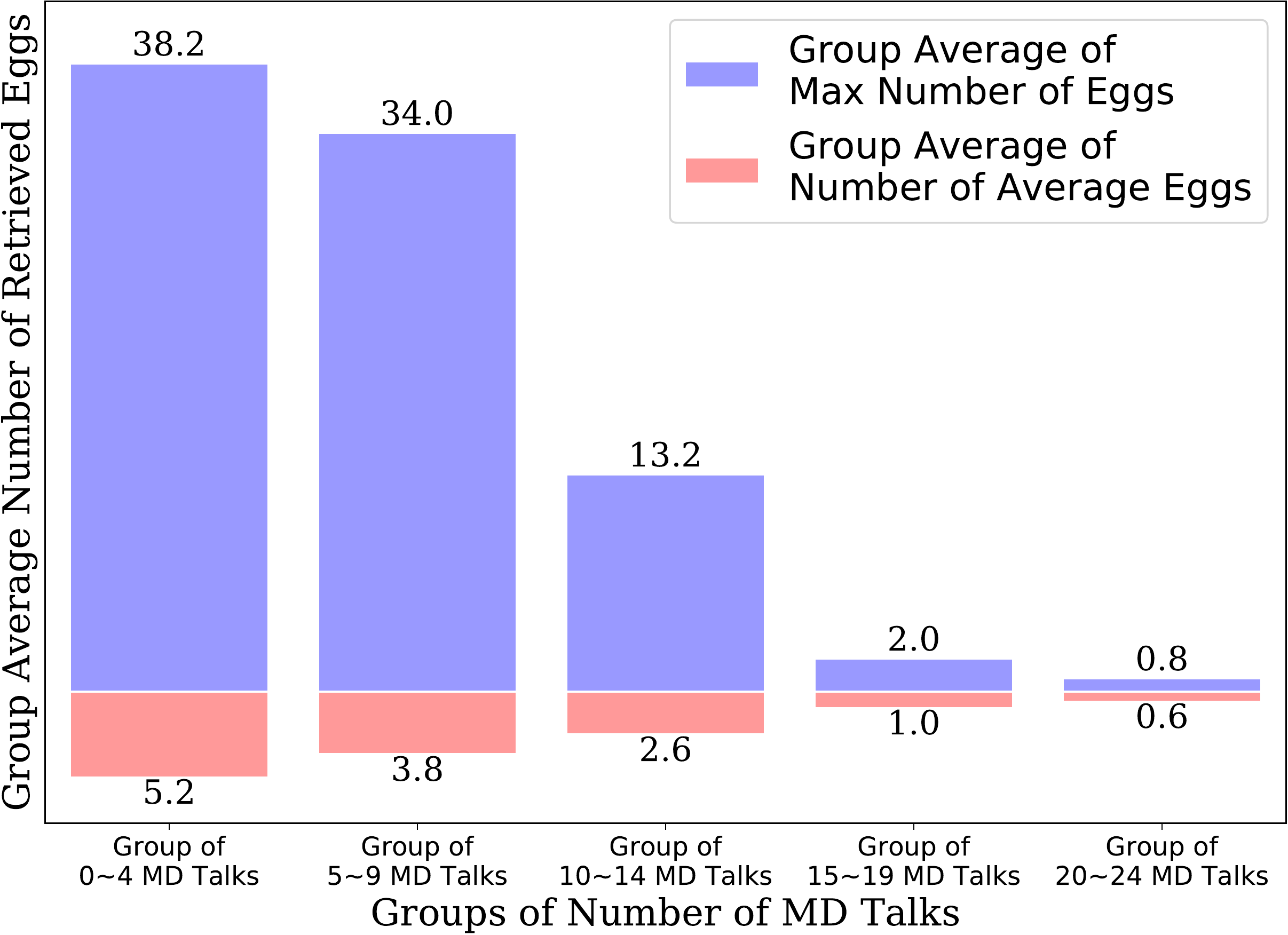}
\caption{Correlation between the MD Talk and Retrieved oocytes.}
\label{md_oocyte}
\end{figure}

\begin{figure}[H]
\centering
\includegraphics[width=78mm]{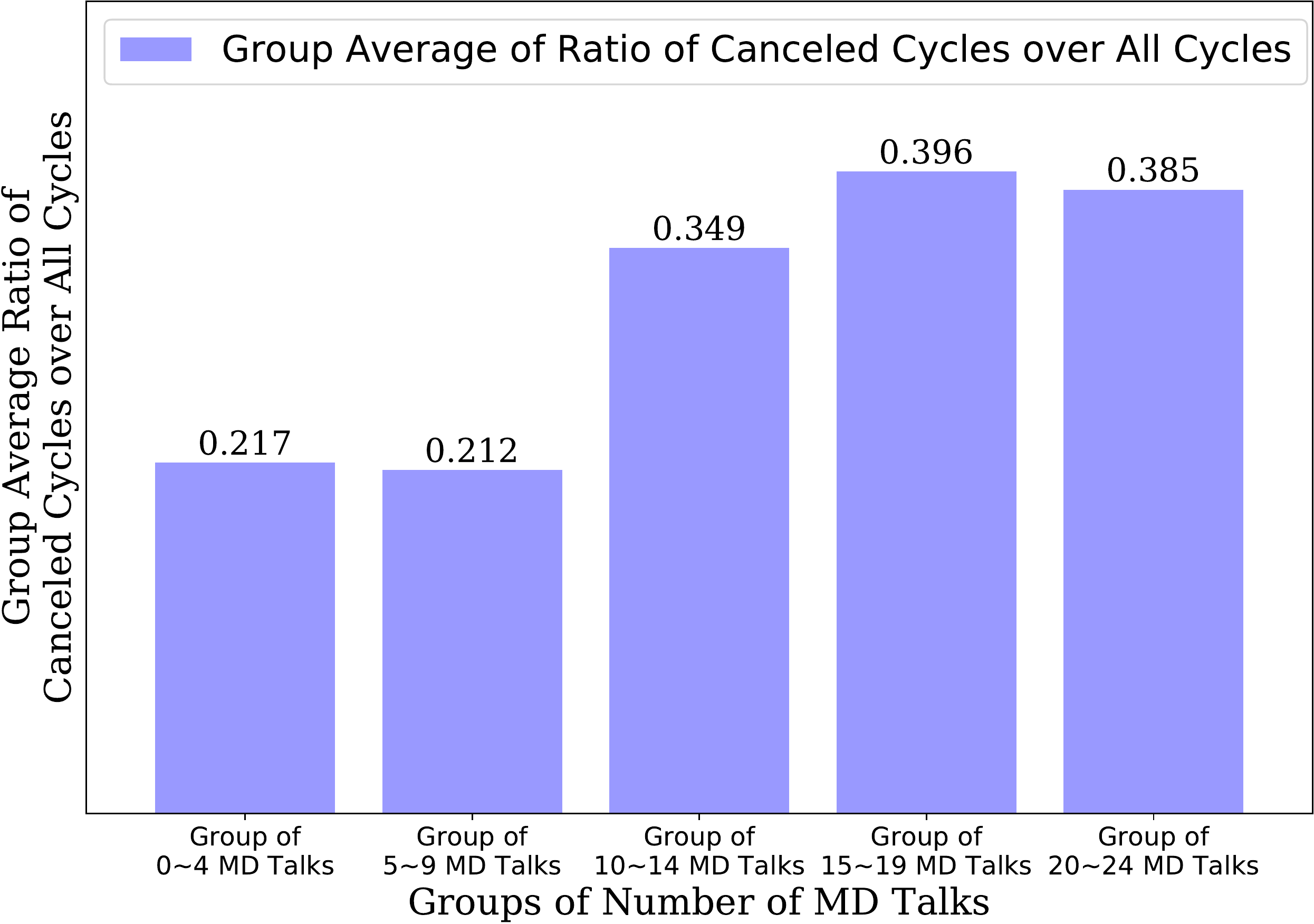}
\caption{Statistics of Number of MD Talks and Number of Cycles.}
\label{md_cycle}
\end{figure}

Prompt MD talk alerts are important for both doctors and patients because on one hand, doctors can give treatments for the condition of patients in time; on the other hand, patients receive high-quality care from the quick response of doctors. So a desirable feature of the IVF engine is to precisely capture patient conditions and then to alert doctors when needed. We thus evaluate the accuracy of the MD talk alerts generated by the IVF engine. Since there are no ground truth for MD talk alerts in the dataset, we turn to investigate the correlation between MD talk alerts and cycle results. The investigation is based on the observation fact that, the more conditions (more MD talk alerts) does one IVF cycle encounter, the worse result (fewer oocytes retrieved) will this cycle attain. Figure \ref{md_oocyte} plots the correlation between the number of MD talk alerts predicted by the IVF engine and the number of retrieved oocytes in the same cycle. With the number of MD talk increasing, the maximum, and the average number of retrieved oocyte decreases. Moreover, Figure \ref{md_cycle} illustrates that, except for the scenario of the number of zero MD talks, the cancellation rate of IVF cycles increase with the number of MD talks increasing. Note that the high cancellation ratio on the scenario of the number of zero MD talks results from personal reasons from patients, such as illnesses and financial issues, rather than medical conditions.

\section{Conclusion and Future Work}
In this paper, we proposed a knowledge-based decision support system for IVF treatments. We developed the IVF treatment protocols with a well organized modular approach and implemented the system (knowledge base, inference engine, data base, user interface, etc) that is seamlessly integrated into the existing EMR system. We evaluated our system with a large clinical dataset. Our experiments show that, this system achieves remarkable performance in the accuracy of ad vice for IVF treatments and medications.

For the future work, we plan to improve the IVF engine by employing some machine learning/deep learning approaches on top of the knowledge-based system to make it more flexible and adapt to individual doctor's preference.

% \bibliographystyle{ieee}
% \bibliography{mybib}

\end{document}